\documentclass[letterpaper]{article} 
\usepackage{aaai2027}  
\usepackage[hyphens]{url}  
\usepackage{graphicx} 
\usepackage{natbib}  
\usepackage{caption} 
\usepackage{algorithm}
\usepackage{algorithmic}
\usepackage{xcolor}
\usepackage{colortbl}

\usepackage{newfloat}
\usepackage[table]{xcolor}
\usepackage{amsfonts}
\usepackage{listings}
\usepackage{multirow}
\usepackage{tabularx}
\usepackage{array}
\DeclareCaptionStyle{ruled}{labelfont=normalfont,labelsep=colon,strut=off} 
\floatstyle{ruled}
\newfloat{listing}{tb}{lst}{}
\floatname{listing}{Listing}

\usepackage{booktabs}

\usepackage{amsmath}

\title{SAGE: Variate-Wise Semantic Augmentation for Vision-Language \\Time Series Forecasting}
\author{
    Haizhao Fan\textsuperscript{\rm 1}, Xinyi Le\textsuperscript{\rm 1} 
}
\affiliations{
    \textsuperscript{\rm 1}Shanghai Jiao Tong University\\
    Shanghai, China\\
    sharp-ro@sjtu.edu.cn, lexinyi@sjtu.edu.cn
}

\begin{document}

\makeatletter
\gdef\copyright@on{}
\makeatother
\nocopyright

\maketitle

\begin{abstract}
Time series forecasting models operate on raw numerical sequences, lacking the semantic knowledge that domain experts implicitly leverage, such as the physical meaning of each variable, its statistical behavior, and its temporal dynamics. Recent efforts to bridge this gap fall into two camps. Some rely on large language models at inference time, which is computationally expensive. Others apply uniform textual prompts at the dataset level, ignoring the heterogeneous semantics across individual variates. We propose \textbf{SAGE} (\textbf{S}eeing and
\textbf{A}ugmenting with \textbf{G}rounded \textbf{E}ncoding), an end-to-end
CLIP-based framework that jointly models temporal, cross-variable, textual, and
visual information. The CLIP text encoder processes frequency-enhanced patches
and variable tokens, while gated residual paths inject variable-specific
descriptions and statistical descriptors. In parallel, the frozen CLIP vision
encoder aligns rendered series with temporal representations through a
training-only contrastive objective. This dual use of CLIP adds complementary
semantic and visual supervision without placing an LLM in the forecasting
loop. Across eight long-term benchmarks and M4, SAGE achieves state-of-the-art
accuracy. Ablations confirm complementary gains from multimodal alignment and
variable-level knowledge.
\end{abstract}


\section{Introduction}

Time series forecasting is a fundamental task in data mining and machine
learning, with wide-ranging applications in energy management, transportation
planning, weather prediction, and financial
analysis~\citep{zhou2020informer,wu2021autoformer}. The advent of
Transformer-based architectures has brought significant
progress~\citep{zhou2022fedformer,liu2024itransformer}, while the provocative
finding that simple linear models can match or surpass many Transformer
variants~\citep{zeng2023transformers} has spurred the community to rethink
what inductive biases truly matter for temporal modeling.

\begin{figure*}[t]
\centering
\includegraphics[width=\textwidth]{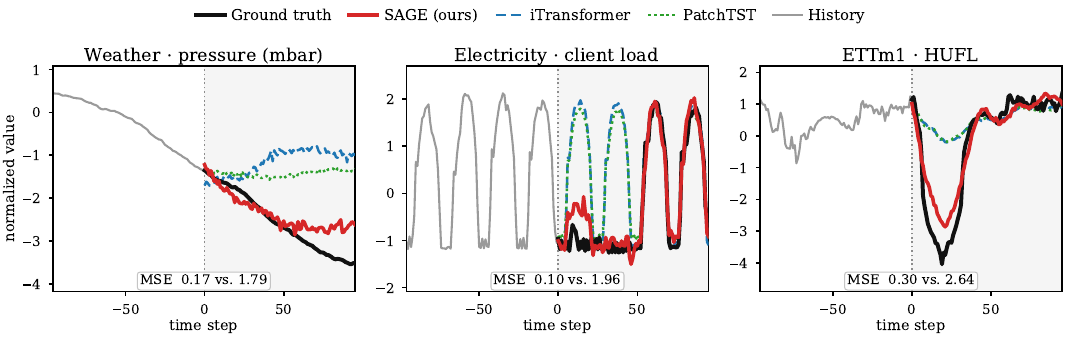}
\caption{Representative 96-step forecasts on Weather, Electricity, and ETTm1.
SAGE better captures regime changes by combining numerical history with
multimodal knowledge. Each inset reports window-level MSE for SAGE and the
strongest displayed baseline.}
\label{fig:teaser}
\end{figure*}

More recently, two parallel trends have reshaped this landscape. The first is
\textbf{large-scale pre-training}: foundation models such as
TimesFM~\citep{das2024decoder}, Chronos~\citep{ansari2024chronos}, and
Time-MoE~\citep{shi2025time} demonstrate that pre-training on billions of time
points yields strong zero-shot and few-shot forecasting capabilities. The
second is \textbf{cross-modal knowledge transfer}: methods such as
GPT4TS~\citep{zhou2023one}, Time-LLM~\citep{jin2024time}, and
TEST~\citep{sun2024test} repurpose pre-trained language models for time series
tasks, exploiting their rich semantic and sequential knowledge. However, the
LLM component in many of these methods can be replaced by simple attention
layers without degrading performance~\citep{tan2024language}, raising questions
about whether language models truly contribute temporal understanding or
merely serve as general-purpose feature extractors.

Meanwhile, an emerging body of work explores \textbf{vision-language models
(VLMs)} for time series. A recent survey~\citep{ni2025harnessing} highlights
the natural compatibility between 2D visual representations and multi-scale
temporal patterns, an insight previously demonstrated by
TimesNet~\citep{wu2022timesnet}. Time-VLM~\citep{zhong2025time} fuses
retrieval, vision, and text modalities for forecasting;
Aurora~\citep{wu2025aurora} builds the first multimodal time series foundation
model; and OccamVTS~\citep{lyu2026occamvts} shows that distilling vision
models to 1\% of their parameters can match full-model forecasting
performance. The CiK benchmark~\citep{williams2024context} further establishes
that textual context can be essential for accurate prediction in
many real-world scenarios.

Despite these advances, a fundamental gap remains: existing multimodal methods
typically use language models as either frozen feature extractors or auxiliary
prompt generators, without fully exploiting the \emph{dual structure} of
vision-language models. In CLIP~\citep{radford2021learning}, the text and
vision encoders are jointly trained to produce aligned representations in a
shared embedding space. This dual structure naturally suits time series. The
text encoder can serve as a sequential backbone, while the vision encoder
provides complementary supervision through rendered time series images.
Moreover, existing knowledge-injection schemes
~\citep{jin2024time,pan2024s} may require expensive LLM inference or reduce
numerical sequences to limited textual representations.
Figure~\ref{fig:teaser} previews the payoff of closing this gap: across real
test windows from three different datasets, SAGE tracks the periodic and
directional dynamics that strong Transformer forecasters collapse toward the
recent mean.

In this paper, we propose SAGE, a multimodal semantic alignment framework for
time series forecasting that addresses these limitations through three key
contributions:

\begin{enumerate}
\item \textbf{End-to-End CLIP Forecasting.} We make CLIP part of the
forecasting architecture rather than an external feature service. Its text
encoder is adapted as the temporal backbone, and its frozen vision encoder
provides bidirectional contrastive supervision during training. This design
achieves higher forecasting accuracy with a compact trainable backbone instead
of relying on a billion-parameter model in the forecasting loop.

\item \textbf{Template-based Knowledge Injection with Variable-wise Gating.} A
lightweight offline pipeline constructs semantic, behavioral, and relational
descriptions from metadata and training-set statistics. An LLM may assist this
preparation step, but it is not called inside training or forecasting. Gated
cross-attention controls knowledge injection independently for each variable,
while a separately gated statistical bypass preserves hard numerical cues.

\item \textbf{Grounding Across Time, Variables, and Modalities.} A
frequency-enhanced patch embedding combines time-domain and spectral evidence.
Cross-variate attention then incorporates dependencies among variables, while
vision-language contrastive alignment regularizes the temporal
representations. Together, these components integrate complementary temporal,
relational, textual, and visual signals in one forecasting model.
\end{enumerate}

\section{Related Work}

\subsection{Multimodal and Knowledge-Enhanced Time Series Forecasting}

Early work demonstrated that pre-trained language models can serve as
effective time series backbones: GPT4TS~\citep{zhou2023one} fine-tunes only
the layer normalization and positional embeddings of a frozen GPT-2, while
LLMTime~\citep{gruver2023large} encodes numerical series as text strings for
zero-shot forecasting. Subsequent methods introduce explicit cross-modal
alignment: Time-LLM~\citep{jin2024time} reprograms input patches into text
prototypes guided by declarative prompts, and TEST~\citep{sun2024test} aligns
time series embeddings with interpretable text prototypes via contrastive
losses. A recurring limitation is that the language component often acts as a
generic feature extractor rather than a genuine source of temporal
knowledge~\citep{tan2024language}.

More recently, the community has moved toward true multimodal fusion.
Time-VLM~\citep{zhong2025time} couples retrieval-, vision-, and text-augmented
learners with frozen VLMs; Aurora~\citep{wu2025aurora} injects text and image
knowledge via modality-guided attention and flow matching for zero-shot
generative forecasting; and VLM4TS~\citep{he2026harnessing} applies VLMs to
zero-shot anomaly detection through visual screening and multimodal
verification. Despite these advances, most existing methods treat the language
model as either a frozen feature extractor or an auxiliary prompt generator,
without fully exploiting both branches of a vision-language model in a unified
training framework.

\subsection{Pre-trained Models for Time Series}

In the supervised regime, Autoformer~\citep{wu2021autoformer} pioneers deep
decomposition with auto-correlation mechanisms; PatchTST~\citep{nie2022time}
establishes a strong patch-based channel-independent baseline;
iTransformer~\citep{liu2024itransformer} treats each variate as a token to
capture multivariate correlations; and TimesNet~\citep{wu2022timesnet}
transforms 1D series into 2D tensors, enabling vision backbones for temporal
modeling. More recent architectures continue to refine this line:
FredFormer~\citep{piao2024fredformer} debiases frequency-domain representations,
DUET~\citep{qiu2025duet} jointly clusters temporal and channel dependencies,
Amplifier~\citep{fei2025amplifier} amplifies energy-scarce components, and
SRSNet~\citep{wu2026enhancing} learns a selective representation space; these are
the strong recent baselines against which we benchmark SAGE. In the foundation
model regime, TimesFM~\citep{das2024decoder}
trains a decoder-only Transformer on 100B real-world time points;
Chronos~\citep{ansari2024chronos} tokenizes values into a discrete vocabulary
for T5-family models; MOMENT~\citep{goswami2024moment} pre-trains on the
Time-series Pile via masked reconstruction; Moirai~\citep{woo2024unified}
introduces any-variate attention trained on 27B observations; and
Time-MoE~\citep{shi2025time} scales to 2.4B parameters via sparse
mixture-of-experts. These efforts demonstrate the effectiveness of scale, yet
they require curating massive time series corpora. An alternative direction,
exemplified by OccamVTS~\citep{lyu2026occamvts}, suggests that pre-trained
vision-language representations already contain transferable temporal
knowledge that can be leveraged without billion-scale pre-training.

\subsection{Contrastive Learning for Time Series}

Contrastive learning has proven effective for learning transferable time
series representations: TS-TCC~\citep{eldele2021time} combines temporal and
contextual contrasting over augmented views; TS2Vec~\citep{yue2022ts2vec}
performs hierarchical contrastive learning at instance and temporal levels;
CoST~\citep{woo2022cost} disentangles seasonal and trend components via time-
and frequency-domain losses; TF-C~\citep{zhang2022self} enforces
time-frequency consistency in a joint embedding space;
Soft-CL~\citep{lee2024soft} mitigates false negatives through soft
assignments; and FACL~\citep{wang2026frequency}
designs frequency-aware augmentations that respect spectral structure. All of
these operate within the time series modality, constructing positive pairs
from augmented views of the same data. A natural extension is cross-modal
contrastive learning, where positive pairs couple temporal representations
with their visual renderings, providing a complementary supervisory signal
beyond intra-modal augmentation.

\section{Methodology}

\begin{figure*}[t]
\centering
\includegraphics[width=\textwidth]{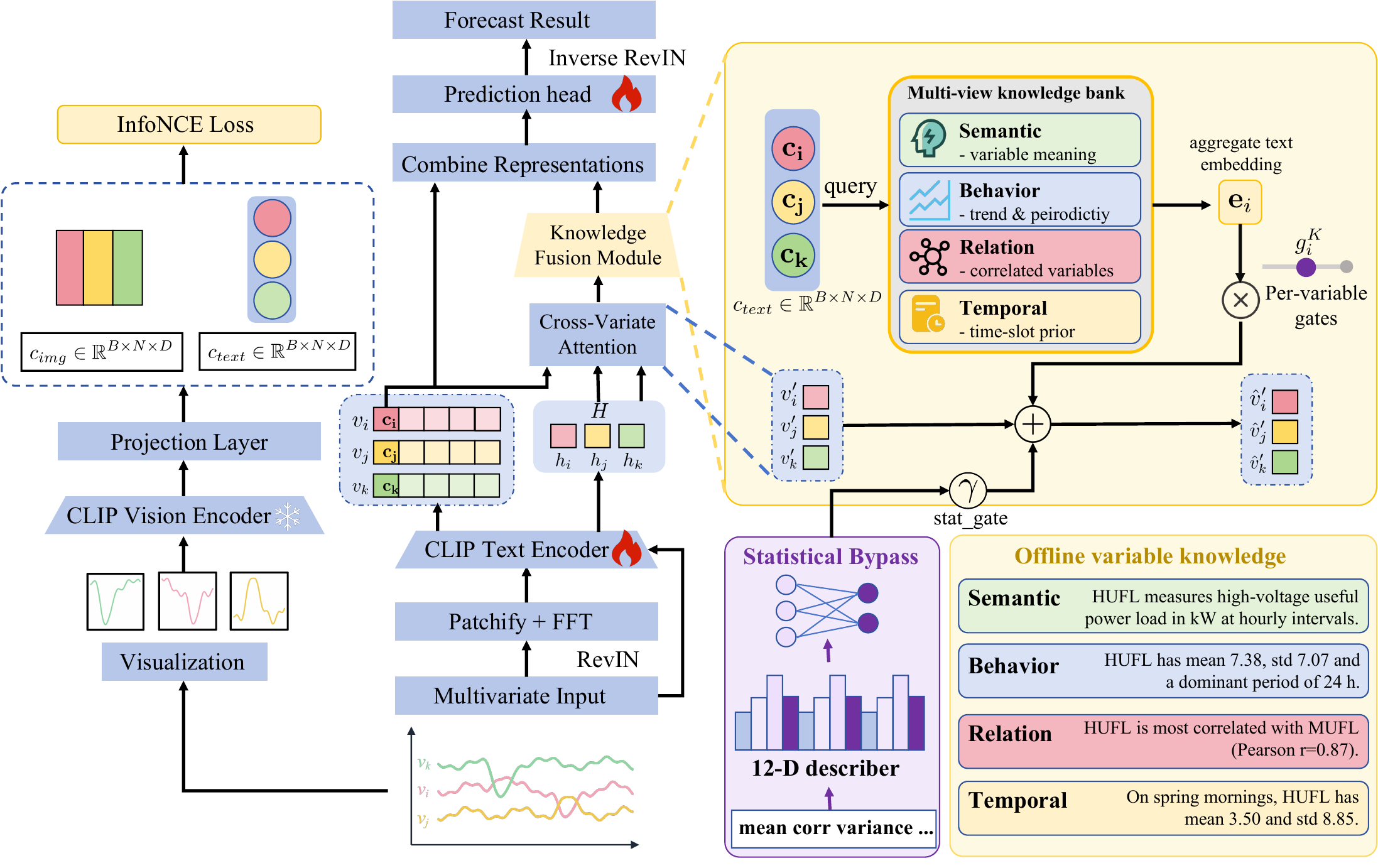}
\caption{Overview of \textbf{SAGE}. Shared CLIP text branches encode frequency-enhanced temporal tokens and cross-variable context. Variable-specific text views use gated attention, while hard statistics follow a separate residual bypass. During training, the frozen CLIP vision encoder provides contrastive supervision. Fused representations are decoded and inverse-normalized. The lower callout traces a real ETTh1/HUFL knowledge example.}
\label{fig:pipeline}
\end{figure*}

We propose a multimodal framework for multivariate time series forecasting. Given an input multivariate time series $\mathbf{X} \in \mathbb{R}^{T \times N}$ with $T$ time steps and $N$ variates, the model outputs a forecast $\hat{\mathbf{Y}} \in \mathbb{R}^{H \times N}$ over a horizon of $H$ steps. As illustrated in Figure~\ref{fig:pipeline}, the normalized input is processed by two cooperating streams that share a single pretrained CLIP text encoder. A \emph{numerical stream} converts each series into patch tokens and variate tokens and passes them through the CLIP text encoder, producing temporal and inter-variate representations that are further enriched with frequency information and with per-variate natural-language priors drawn from an offline textual knowledge bank. A \emph{visual stream}, active only for low-dimensional datasets, renders each series as an image, encodes it with the frozen CLIP vision encoder, and contrastively aligns it with the temporal representation during training. A forecast head then fuses these representations and de-normalizes them to produce the prediction. Concretely, the framework comprises five components that we describe in turn: a Frequency-Enhanced Language Module, an Inter-Variate Dependency Module, a Multi-View Textual Semantic Fusion Module, a Vision-Language Contrastive Alignment Module, and a Forecast Generator.

\subsection{Frequency-Enhanced Language Module}

This module treats time series patches as tokens for the pretrained CLIP text
Transformer.

We first apply reversible instance normalization to mitigate distribution shift.
For variable $i$, RevIN stores its mean $\mu^{(i)}$ and standard deviation
$\sigma^{(i)}$ for output de-normalization. We omit the batch index in
per-variable equations for clarity. The normalized sequence
$\mathbf{x}^{(i)}\in\mathbb{R}^{T}$ is divided into $L_p$ overlapping patches
of length $P$ and stride $S$. Let $\mathbf{p}^{(i)}_j\in\mathbb{R}^{P}$ denote
patch $j$. A learnable tokenizer $\mathbf{W}_{\mathrm{tok}}$ maps it to the
shared embedding width $D$:
\[
\mathbf{z}^{(i)}_j=\mathbf{W}_{\mathrm{tok}}\mathbf{p}^{(i)}_j.
\]

Time-domain tokens alone may inadequately represent periodic patterns. We apply
a Hann-windowed FFT to every patch and project its real and imaginary
components into a frequency token
$\mathbf{f}^{(i)}_j\in\mathbb{R}^{D}$. The time token is the query, and the
frequency token provides the key and value. A learnable scalar $\alpha$,
initialized at zero, controls this fusion:

\[
\mathbf{z}^{(i)}_j \leftarrow \mathbf{z}^{(i)}_j
+\alpha\,\operatorname{CrossAttn}
\bigl(\mathbf{z}^{(i)}_j,\mathbf{f}^{(i)}_j,\mathbf{f}^{(i)}_j\bigr),
\]
where $\operatorname{CrossAttn}(Q,K,V)$ denotes cross-attention with query
$Q$, key $K$, and value $V$.

A learnable [CLS] token is prepended, and learnable positional embeddings are
added to the patch sequence. The sequence then enters the CLIP text Transformer
without using its word embedding layer. We replace the original positional
embeddings with embeddings matched to the patch sequence. The Transformer
blocks and final LayerNorm are fine-tuned at a reduced learning rate, while
unused projection layers remain frozen. The encoder outputs patch
representations
$\mathbf{R}\in\mathbb{R}^{B\times N\times(L_p+1)\times D}$ and temporal
summaries
$\mathbf{c}_{\text{text}}\in\mathbb{R}^{B\times N\times D}$ from the [CLS]
positions.

\subsection{Cross-Variate Context Modeling Module}

The temporal encoder processes each variable independently, but multivariate
forecasting also requires dependencies across variables.

Each full normalized sequence $\mathbf{x}^{(i)}$ is mapped to the shared
embedding width by a linear tokenizer and augmented with a learnable variable
identifier. The $N$ resulting tokens pass through the same CLIP text encoder
used above. Its output
$\mathbf{H}\in\mathbb{R}^{B\times N\times D}$ contains contextual
representations of all variables. Thus, the shared encoder processes patch
sequences to capture temporal structure and variable sequences to capture
cross-variable structure.

The temporal summaries query $\mathbf{H}$ through cross-attention. Layer
normalization $\operatorname{LN}$ and a feed-forward network
$\operatorname{FFN}$ form residual updates:

\[
\mathbf{v}'=
\operatorname{LN}\!\left(
\mathbf{c}_{\text{text}}+
\operatorname{CrossAttn}\bigl(
\operatorname{LN}(\mathbf{c}_{\text{text}}),
\operatorname{LN}(\mathbf{H}),
\operatorname{LN}(\mathbf{H})
\bigr)\right),
\]
\[
\mathbf{v}'=
\operatorname{LN}\!\left(
\mathbf{v}'+\operatorname{FFN}(\mathbf{v}')
\right).
\]

The result
$\mathbf{v}'\in\mathbb{R}^{B\times N\times D}$ combines each
variable's temporal summary with information selected from all variables.

\subsection{Multi-View Textual Semantic Fusion Module}

This module is the central contribution of our work. It injects external semantic knowledge into the forecasting pipeline through three complementary mechanisms.

\begin{table}[!t]
\centering
\small
\caption{Offline variable-specific knowledge resources. A real ETTh1/HUFL
example is visualized in Figure~\ref{fig:pipeline}.}
\label{tab:text_taxonomy}
\setlength{\tabcolsep}{3pt}
\renewcommand{\arraystretch}{1.10}
\begin{tabularx}{\columnwidth}{
    @{}l
    >{\raggedright\arraybackslash}X
    l@{}
}
\toprule
Source & Information & Fusion \\
\midrule
Semantic   & Meaning, unit, sampling context
           & Attention \\
Behavioral & Smoothness, volatility, periodicity
           & Attention \\
Relational & Cross-variate correlations
           & Attention \\
Temporal   & Seasonal and time-of-day patterns
           & Extra view \\
Statistics & Distribution, trend, spectrum
           & MLP bypass \\
\bottomrule
\end{tabularx}
\end{table}

For each variable, we construct descriptions from domain metadata and
training-split statistics. Table~\ref{tab:text_taxonomy} summarizes the
knowledge sources, while Figure~\ref{fig:pipeline} shows representative HUFL
examples. The CLIP text encoder maps the descriptions to the shared embedding
space. We also compute a
12-dimensional statistical vector
$\mathbf{s}^{(i)}\in\mathbb{R}^{12}$ for variable $i$. All descriptions,
embeddings, and statistical vectors are cached before model training.

\paragraph{Template-based Construction Pipeline.}
Each human-readable template combines domain metadata with data-driven
features. The metadata include the variable name, unit, and domain role from
the dataset documentation. Training-set features include distributional
statistics, trend slope, dominant FFT period, and Pearson correlations with
other variables. The templates may be authored or refined with optional LLM
assistance during offline preparation. The relational view names the
most correlated variables, restoring context that a channel-independent
encoder would otherwise omit. Table~\ref{tab:ablation} evaluates five
enhancement modes that control which feature groups are verbalized.

Let $\mathbf{E}_{\mathrm{mv}}\in\mathbb{R}^{N\times K\times D}$ contain $K$
cached views for each variable. Variable-wise cross-attention uses
$\mathbf{c}_{\text{text}}^{(b,i)}$ as its query and produces an aggregated
embedding $\mathbf{e}^{(b,i)}\in\mathbb{R}^{D}$. When time-dependent text is
available, the input timestamp selects the corresponding slot embedding and
appends it as an additional view.

The aggregated embedding is fused with the cross-variate representation through
a learnable gate $g^{(i)}$ for variable $i$:

\[
\begin{aligned}
\mathbf{v}_{\text{text}}^{\prime\,(b,i)}
&={}\mathbf{v}^{\prime\,(b,i)} \\
&+g^{(i)}\operatorname{CrossAttn}\!\left(
\operatorname{LN}(\mathbf{v}^{\prime\,(b,i)}),
\operatorname{LN}(\mathbf{e}^{(b,i)}),
\mathbf{e}^{(b,i)}
\right)
\end{aligned}
\]

The gate is initialized to a small positive value, allowing every variable to
learn its own reliance on textual priors. Statistical features bypass
attention. A two-layer network
$\operatorname{MLP}_{\mathrm{stat}}:\mathbb{R}^{12}\rightarrow\mathbb{R}^{D}$
and a separate learnable gate $\gamma_{\mathrm{stat}}$ inject them after text
attention:
\[
\hat{\mathbf{v}}^{\prime\,(b,i)}
=\mathbf{v}_{\text{text}}^{\prime\,(b,i)}
+\gamma_{\mathrm{stat}}\,
\operatorname{MLP}_{\mathrm{stat}}\!\left(\mathbf{s}^{(i)}\right).
\]
This bypass complements soft textual semantics with hard numerical evidence
without consuming attention capacity.

We also align temporal and aggregated textual representations:
\[
\mathcal{L}_{\text{text}}
=1-\frac{1}{BN}\sum_{b=1}^{B}\sum_{i=1}^{N}
\cos\!\left(
\mathbf{c}_{\text{text}}^{(b,i)},\mathbf{e}^{(b,i)}
\right).
\]

\subsection{Vision-Language Contrastive Alignment Module}

When $N\leq 50$, an auxiliary vision-language objective provides additional
training supervision. Each normalized variable sequence is rendered as a
color-coded line chart, encoded by the frozen CLIP vision encoder, and mapped
to the shared width by a learnable projection
$\mathbf{W}_{\mathrm{proj}}$:

\[
\mathbf{c}_{\text{img}}^{(b,i)}
=\mathbf{W}_{\mathrm{proj}}\,
\operatorname{CLIP\text{-}ViT}\!\left(
\operatorname{Render}(\mathbf{x}^{(b,i)})
\right),
\quad
\mathbf{c}_{\text{img}}^{(b,i)}\in\mathbb{R}^{D}
\]

Alignment is performed \emph{independently for each variate}. For variate $i$,
let $\mathbf{S}_{\text{img}}^{(i)},\mathbf{S}_{\text{text}}^{(i)}\in\mathbb{R}^{B\times D}$
stack the L2-normalized visual and temporal summaries of the $B$ samples in the
batch. A bidirectional InfoNCE loss matches each sample to its own rendering,
using the remaining $B-1$ same-variate samples as in-batch negatives, and the
result is averaged over the $N$ variates:

\[
\begin{aligned}
\mathcal{L}_{\text{align}}
=\frac{1}{N}\sum_{i=1}^{N}\frac{1}{2}\Big[
&\operatorname{CE}\!\left(
\tau\,\mathbf{S}_{\text{img}}^{(i)}\mathbf{S}_{\text{text}}^{(i)\top},\mathbf{I}_{B}
\right)\\
&+\operatorname{CE}\!\left(
\tau\,\mathbf{S}_{\text{text}}^{(i)}\mathbf{S}_{\text{img}}^{(i)\top},\mathbf{I}_{B}
\right)
\Big],
\end{aligned}
\]

where $\mathbf{I}_{B}$ denotes the batch-level matching-pair targets (the identity
assignment over the $B$ samples) and $\tau=\exp(s)$ is the learnable logit scale.

A curriculum schedule gradually increases the concentration of hard negatives.
This module is active only during training. For $N>50$, rendering hundreds of
variable-wise images is impractical, so the vision loss is disabled and textual
knowledge fusion remains active.

\subsection{Forecast Generator and Training Objective}

The augmented representation $\hat{\mathbf{v}}'$ is added to the final
position of $\mathbf{R}$. This operation injects cross-variable and external
knowledge without replacing earlier patch representations. The resulting token
sequence is flattened and mapped to $T_{\mathrm{pred}}$ forecast steps by a
two-layer MLP with GELU activation. RevIN then restores the original scale.

For $N>50$, the full variable-wise patch path is replaced by a lightweight
decoder for computational practicality. Cross-variate context modeling and
textual semantic fusion remain active, and a linear layer maps each augmented
variable representation to $T_{\mathrm{pred}}$ outputs. When time-dependent
descriptions are available, the selected slot embedding supplies an additional
gated correction at each forecast step.

The total objective combines forecast MSE with the two alignment losses:

\[
\mathcal{L}
=\mathcal{L}_{\text{MSE}}
+\lambda_{\text{align}}\mathcal{L}_{\text{align}}
+\lambda_{\text{text}}\mathcal{L}_{\text{text}},
\]
where $\mathcal{L}_{\text{MSE}}$ is the error between
$\hat{\mathbf{Y}}$ and the ground-truth future sequence.

The coefficients $\lambda_{\text{align}}$ and $\lambda_{\text{text}}$ weight
vision-language and text alignment, respectively. The CLIP text encoder is
optimized at a reduced learning rate to preserve pretrained knowledge while
allowing task-specific adaptation.

\section{Experiments}

\subsection{Setup}

\paragraph{Datasets.}

We evaluate SAGE on 8 widely-used long-term forecasting benchmarks and the M4 short-term forecasting competition.
The long-term benchmarks include: ETTh1, ETTh2, ETTm1, ETTm2 (electricity transformer temperature, $N{=}7$)~\citep{zhou2020informer},
Electricity (ECL, $N{=}321$), Traffic ($N{=}862$), Weather ($N{=}21$)~\citep{wu2021autoformer},
and Exchange ($N{=}8$)~\citep{lai2017modeling}.
For long-term forecasting, we use a fixed lookback window $T{=}96$ and
prediction horizons
$T_{\mathrm{pred}}\in\{96,192,336,720\}$.
For short-term forecasting, we evaluate on the M4 dataset~\citep{makridakis2020m4} comprising 100,000 time series across six frequencies (Yearly, Quarterly, Monthly, Weekly, Daily, Hourly).

\begin{table}[t]
\centering
\small
\caption{Long-term benchmark characteristics and SAGE pathways.
All datasets use $T=96$ and prediction horizons in
$\{96,192,336,720\}$.}
\label{tab:dataset_overview}
\setlength{\tabcolsep}{3.2pt}
\renewcommand{\arraystretch}{1.10}
\begin{tabular}{@{}lrrll@{}}
\toprule
Dataset & $N$ & Observations & Interval & Pathway \\
\midrule
ETTh1    & 7   & 17,420 & 1 hour     & Full        \\
ETTh2    & 7   & 17,420 & 1 hour     & Full        \\
ETTm1    & 7   & 69,680 & 15 minutes & Full        \\
ETTm2    & 7   & 69,680 & 15 minutes & Full        \\
ECL      & 321 & 26,304 & 1 hour     & Lightweight \\
Traffic  & 862 & 17,544 & 1 hour     & Lightweight \\
Weather  & 21  & 52,696 & 10 minutes & Full        \\
Exchange & 8   & 7,588  & 1 day      & Full        \\
\bottomrule
\end{tabular}
\end{table}

Table~\ref{tab:dataset_overview} highlights the diversity of the evaluation
suite. The four ETT datasets measure electricity-transformer behavior at
hourly and 15-minute resolutions; ECL and Traffic stress scalability with
hundreds of variables; Weather provides densely sampled meteorological
signals; and Exchange contains daily financial series. 

\paragraph{Metrics.}
Long-term forecasting is evaluated with Mean Squared Error (MSE) and Mean Absolute Error (MAE).
Short-term forecasting follows the M4 competition protocol with SMAPE, MASE, and Overall Weighted Average (OWA).

\begin{table*}[!t]
\caption{Dataset-averaged MSE / MAE comparison across prediction horizons $\{96, 192, 336, 720\}$ with lookback $T{=}96$.
\textbf{Bold}: best; \underline{underline}: second best.}
\label{tab:main_results}
\centering
\resizebox{\textwidth}{!}{
\begin{tabular}{c|cc|cc|cc|cc|cc|cc|cc|cc|cc}
\toprule
\multirow{2}{*}{Dataset}
& \multicolumn{2}{c}{DLinear} & \multicolumn{2}{c}{TimesNet} & \multicolumn{2}{c}{PatchTST} & \multicolumn{2}{c}{FredFormer} & \multicolumn{2}{c}{iTransformer} & \multicolumn{2}{c}{Amplifier} & \multicolumn{2}{c}{DUET} & \multicolumn{2}{c}{SRSNet} & \multicolumn{2}{c}{SAGE} \\
& \multicolumn{2}{c}{\citeyearpar{zeng2023transformers}} & \multicolumn{2}{c}{\citeyearpar{wu2022timesnet}} & \multicolumn{2}{c}{\citeyearpar{nie2022time}} & \multicolumn{2}{c}{\citeyearpar{piao2024fredformer}} & \multicolumn{2}{c}{\citeyearpar{liu2024itransformer}} & \multicolumn{2}{c}{\citeyearpar{fei2025amplifier}} & \multicolumn{2}{c}{\citeyearpar{qiu2025duet}} & \multicolumn{2}{c}{\citeyearpar{wu2026enhancing}} & \multicolumn{2}{c}{Ours} \\
\cmidrule(lr){2-3}\cmidrule(lr){4-5}\cmidrule(lr){6-7}\cmidrule(lr){8-9}\cmidrule(lr){10-11}\cmidrule(lr){12-13}\cmidrule(lr){14-15}\cmidrule(lr){16-17}\cmidrule(lr){18-19}
& \scalebox{0.85}{MSE} & \scalebox{0.85}{MAE} & \scalebox{0.85}{MSE} & \scalebox{0.85}{MAE} & \scalebox{0.85}{MSE} & \scalebox{0.85}{MAE} & \scalebox{0.85}{MSE} & \scalebox{0.85}{MAE} & \scalebox{0.85}{MSE} & \scalebox{0.85}{MAE} & \scalebox{0.85}{MSE} & \scalebox{0.85}{MAE} & \scalebox{0.85}{MSE} & \scalebox{0.85}{MAE} & \scalebox{0.85}{MSE} & \scalebox{0.85}{MAE} & \scalebox{0.85}{MSE} & \scalebox{0.85}{MAE} \\
\midrule
ETTh1 & 0.456 & 0.449 & 0.458 & 0.450 & 0.469 & 0.450 & 0.449 & 0.442 & 0.454 & 0.442 & 0.449 & 0.442 & 0.443 & \underline{0.438} & \underline{0.442} & \textbf{0.437} & \textbf{0.440} & 0.447 \\
ETTh2 & 0.559 & 0.515 & 0.414 & 0.421 & 0.387 & \underline{0.404} & 0.378 & 0.405 & 0.383 & \underline{0.404} & 0.389 & 0.405 & \underline{0.376} & 0.405 & \underline{0.376} & 0.406 & \textbf{0.373} & \textbf{0.403} \\
ETTm1 & 0.403 & 0.406 & 0.400 & 0.407 & 0.387 & \textbf{0.400} & \underline{0.385} & \underline{0.402} & 0.407 & 0.412 & \underline{0.385} & \underline{0.402} & 0.390 & 0.403 & \underline{0.385} & \underline{0.402} & \textbf{0.384} & \textbf{0.400} \\
ETTm2 & 0.350 & 0.377 & 0.291 & 0.338 & \underline{0.281} & \underline{0.332} & \underline{0.281} & \underline{0.332} & 0.288 & 0.337 & \textbf{0.280} & \textbf{0.331} & \textbf{0.280} & \textbf{0.331} & 0.284 & 0.334 & 0.283 & 0.334 \\
ECL & 0.212 & 0.295 & 0.192 & 0.289 & 0.205 & 0.292 & 0.178 & 0.271 & 0.178 & 0.274 & 0.173 & 0.267 & \underline{0.172} & \underline{0.266} & 0.188 & 0.281 & \textbf{0.165} & \textbf{0.260} \\
Traffic & 0.625 & 0.383 & 0.620 & 0.395 & 0.481 & 0.363 & 0.434 & 0.285 & \underline{0.432} & \underline{0.282} & 0.485 & 0.324 & 0.451 & 0.297 & 0.495 & 0.327 & \textbf{0.430} & \textbf{0.278} \\
Weather & 0.265 & 0.287 & 0.259 & 0.285 & 0.259 & 0.285 & 0.256 & 0.279 & 0.258 & 0.283 & \underline{0.247} & \underline{0.276} & 0.251 & 0.278 & 0.249 & 0.277 & \textbf{0.243} & \textbf{0.271} \\
Exchange & 0.353 & 0.411 & 0.416 & 0.445 & 0.367 & 0.420 & 0.374 & 0.408 & 0.360 & 0.410 & 0.361 & 0.402 & \underline{0.341} & 0.399 & 0.343 & \underline{0.395} & \textbf{0.332} & \textbf{0.393} \\
\midrule
\textbf{1st Count} & 0 & 0 & 0 & 0 & 0 & 1 & 0 & 0 & 0 & 0 & 1 & 1 & 1 & 1 & 0 & 1 & 7 & 6 \\
\bottomrule
\end{tabular}
}
\end{table*}

\paragraph{Implementation.}
SAGE fine-tunes the CLIP ViT-B/32 text encoder and keeps its vision encoder
frozen. RevIN-normalized series use frequency-enhanced patches of length 16
and stride 8~\citep{kim2021reversible}; offline descriptions are fused through
per-variable gates. For $N{>}50$, a lightweight path bypasses patching.
Training uses Adam, cosine scheduling, a reduced learning rate for the CLIP
text encoder, and bidirectional InfoNCE alignment. We evaluate 651
configurations across all datasets, horizons, and five text-enhancement modes.

\paragraph{Environment and Computational Cost.}
All experiments are conducted on an Ubuntu 22.04 server with AMD EPYC 9654
processors, 768\,GB RAM, and four NVIDIA A800 GPUs. Each run uses one GPU. The
software stack comprises Python 3.11, PyTorch 2.5.1, and CUDA 12.1. Training
cost depends on dataset dimensionality and the active pathway. One epoch takes
roughly 17\,s on ETTh1, 21\,s on Exchange, 130\,s on ECL, 250\,s on Weather,
and 230--290\,s on Traffic. Early stopping typically yields convergence within
10 epochs, ranging from under 3 minutes on ETT datasets to under 45 minutes on
Traffic. The complete 651-run sweep consumed approximately 1{,}000 single-GPU
hours and was parallelized across the four GPUs. Offline text construction
takes only a few seconds per dataset and is amortized across all runs.

\subsection{Main Results}

As shown in Table~\ref{tab:main_results}, SAGE obtains the best average MSE on
7 of 8 datasets and the best average MAE on 6 of 8. Its overall average MSE is
0.331, a 4.1\% reduction from the 0.345 achieved by
iTransformer~\citep{liu2024itransformer}.

Amplifier~\citep{fei2025amplifier} and DUET~\citep{qiu2025duet} remain strongest
on ETTm2, each reaching 0.280 MSE and 0.331 MAE. SRSNet
~\citep{wu2026enhancing} obtains the best MAE on ETTh1. Across the full table,
SAGE maintains an overall average MSE of 0.331 and MAE of 0.348 with the fixed
lookback $T{=}96$. The evaluated datasets range from 8 to 862 variables, showing
that the gains extend across very different dimensionalities. Text resources
are prepared offline, so no LLM is invoked in the training or forecasting
loop. \textbf{The appendix provides MSE and MAE for
all 32 dataset-horizon combinations}.

\begin{table}[!b]
\centering
\small
\caption{M4 short-term forecasting. Lower is better; bold and underline mark
the best and second-best results.}
\label{tab:m4}
\setlength{\tabcolsep}{5pt}
\begin{tabular}{lccc}
\toprule
Method & SMAPE & MASE & OWA \\
\midrule
TimesNet          & 11.800 & \underline{1.591} & \underline{0.851} \\
N-HiTS            & \underline{11.936} & 1.610 & 0.861 \\
PatchTST           & 12.072 & 1.622 & 0.869 \\
FEDformer          & 12.614 & 1.732 & 0.918 \\
iTransformer       & 12.792 & 1.756 & 0.931 \\
\midrule
\textbf{SAGE} & \textbf{11.593} & \textbf{1.556} & \textbf{0.834} \\
\bottomrule
\end{tabular}
\end{table}

On the M4 short-term benchmark in Table~\ref{tab:m4}, SAGE leads all three
metrics with an OWA of 0.834. TimesNet and N-HiTS obtain OWA values of 0.851 and
0.861, respectively. The result extends SAGE beyond long-horizon forecasting:
frequency-aware text descriptions complement patch representations at short
horizons as well.

\subsection{Text Enhancement Analysis}

We conduct a systematic ablation over 5 text enhancement modes across all 8
datasets and 4 prediction horizons. For each dataset-horizon pair, all architectural and text-mode hyperparameters are selected using validation MSE. The selected configuration is then evaluated once on the held-out test split.
The 5 modes progressively enrich the text description.
\textbf{+Stat} adds mean, variance, skewness, and kurtosis.
\textbf{+Dyn} adds moving averages and trend slopes.
\textbf{+Stat+Dyn}, abbreviated as +S{+}D, combines both groups.
\textbf{+Dyn+Freq} adds FFT-derived spectral features to dynamic descriptions.
\textbf{+Full} includes statistical, dynamic, spectral, and cross-variable
features.

\begin{table}[!t]
\centering
\small
\caption{Average text-enhancement gains. ``Mode'' is the best enhancement
family and $\Delta$ is the relative MSE reduction.}
\label{tab:ablation}
\setlength{\tabcolsep}{4pt}
\begin{tabular}{lrcccr}
\toprule
Dataset & $N$ & No Text & +Text & Mode & $\Delta$(\%)$\uparrow$ \\
\midrule
ETTh1     &   7 & 0.450 & 0.440 & +Stat  & 2.2 \\
ETTh2     &   7 & 0.381 & 0.373 & +Dyn   & 2.1 \\
ETTm1     &   7 & 0.395 & 0.384 & +Stat  & 2.8 \\
ETTm2     &   7 & 0.294 & 0.283 & +Stat  & 3.7 \\
ECL       & 321 & 0.167 & 0.165 & +Stat  & 1.2 \\
Traffic   & 862 & 0.434 & 0.430 & +Dyn   & 0.9 \\
Weather   &  21 & 0.253 & 0.243 & +Stat  & 3.9 \\
Exchange  &   8 & 0.354 & 0.332 & +S{+}D & 6.3 \\
\midrule
\textbf{Average} & & & & & \textbf{2.8} \\
\bottomrule
\end{tabular}
\end{table}

As shown in Table~\ref{tab:ablation}, text enhancement improves forecasting in 31 of 32 dataset-horizon combinations, with an average MSE reduction of 2.8\%.
Among the modes, +Stat is the most reliable (78.1\% win rate across all experiments, selected as the best mode in 45.2\% of combinations), followed by +Dyn (22.6\%).
The combined mode +Stat+Dyn achieves the highest peak improvement (6.3\% on Exchange), suggesting that statistical and dynamic features provide complementary information.
Interestingly, modes incorporating frequency features (+Dyn+Freq, +Full) are less consistently beneficial, likely because the frequency-enhanced patch embedding already captures spectral structure.

Text gains tend to be larger on lower-dimensional datasets, although the relationship is not strictly monotonic.(Table~\ref{tab:ablation}, column $N$ vs.\ $\Delta$).
Exchange ($N{=}8$) achieves 6.3\% improvement, Weather ($N{=}21$) achieves 3.9\%, ECL ($N{=}321$) achieves 1.2\%, and Traffic ($N{=}862$) achieves 0.9\%.
This pattern reflects two complementary factors:
(1)~the per-variate gate can more precisely modulate text absorption when the number of channels is small,
and (2)~low-dimensional datasets have fewer cross-channel redundancies, making external text knowledge a proportionally larger information source.
The per-variable gates can suppress unhelpful textual information, which limits
degradation on high-dimensional datasets.

\subsection{Vision Enhancement Analysis}

We isolate the contribution of visual supervision by
retraining SAGE with $\lambda_{\mathrm{align}}=0$, while
keeping the model architecture, textual knowledge, optimization
settings, and all other loss terms unchanged. Because the visual
branch is enabled only when $N\leq 50$, the ablation is
conducted on the six eligible datasets. Table
\ref{tab:ablation2} reports the test MSE averaged over
the four prediction horizons.

Vision--temporal alignment improves forecasting on all six
datasets, reducing the average MSE from 0.348 to 0.343, or
1.6\% relatively. The largest gain is obtained on ETTm2 at
2.5\%, followed by ETTh1 and ETTm1 at 1.9\% and 1.8\%.
The ETT and Exchange series generally exhibit visually
identifiable trends and periodic structures, allowing the
rendered views to provide shape information complementary to
the numerical encoder. These consistent gains indicate that
the frozen CLIP vision encoder acts as an effective contrastive
regularizer, encouraging temporal representations to preserve
global trends, periodic geometry, and regime changes.

The smaller 0.4\% gain on Weather can be related to its
dataset characteristics. Weather contains 21 heterogeneous and
strongly coupled indicators sampled every 10 minutes
\citep{wu2021autoformer}; thus, a lookback of 96 covers only
16 hours rather than a complete daily cycle. Moreover, rendering
each variable independently omits interactions among
temperature, humidity, pressure, wind, and radiation, while
similar meteorological channels may act as false negatives in
the identity-based InfoNCE objective. Consequently, the visual
branch provides less additional information beyond the
frequency-enhanced and cross-variate representations.

Importantly, the visual branch is used only during training and
does not provide image features directly to the forecast head.
Therefore, the observed improvements come from representation
regularization rather than additional inference inputs, and
introduce no visual computation at forecasting time.

\begin{table}[!t]
\centering
\small
\caption{Average vision-enhancement gains. $\Delta$ is the relative MSE reduction.}
\label{tab:ablation2}
\setlength{\tabcolsep}{4pt}
\begin{tabular}{lrcccr}
\toprule
Dataset & $N$ & No Vision & +Vision  & $\Delta$(\%)$\uparrow$ \\
\midrule
ETTh1     &   7 & 0.448 & 0.440  & 1.9 \\
ETTh2     &   7 & 0.378 & 0.373  & 1.3 \\
ETTm1     &   7 & 0.391 & 0.384  & 1.8 \\
ETTm2     &   7 & 0.290 & 0.283  & 2.5 \\
Weather   &  21 & 0.244 & 0.243  & 0.4 \\
Exchange  &   8 & 0.337 & 0.332  & 1.5 \\
\midrule
\textbf{Average} & & & & \textbf{1.6} \\
\bottomrule
\end{tabular}
\end{table}

\section{Conclusion}

We present SAGE, a vision-language framework that augments time series forecasting with template-based textual knowledge.
SAGE repurposes a pre-trained CLIP ViT-B/32 as a dual-use backbone, freezing the vision encoder as a contrastive anchor while fine-tuning the text encoder. On this backbone, it injects statistical, dynamic, and spectral descriptions into the forecasting pipeline, with no language model in the forecasting loop.
Three design choices are central to the framework:
(1)~a dual-use CLIP
backbone, with a trainable text encoder and a frozen vision encoder for
training-only contrastive supervision;
(2)~variable-specific multi-view knowledge
injection with independent gates and a statistical bypass;
and (3)~frequency-enhanced temporal encoding combined with cross-variable context.

SAGE obtains the best average MSE on 7 of 8 long-term datasets, the best
average MAE on 6 of 8, and the best M4 OWA of 0.834. Text enhancement improves
31 of 32 dataset-horizon settings. These results show that aligned temporal,
relational, textual, and visual evidence can improve a compact forecasting
backbone.

\paragraph{Limitations and Future Work.}
Several directions remain open.
First, the current text generation relies on hand-crafted templates; learning to compose descriptions end-to-end, potentially via a small language model distilled from domain experts, could further improve text quality and domain adaptability.
Second, the text benefit diminishes on high-dimensional datasets, suggesting that the per-variate gate may benefit from group-wise or hierarchical designs that share text representations across correlated channels.
Third, the current descriptions are constructed from dataset metadata and training-split statistics, without incorporating external event or ontology information could unlock additional gains, particularly for datasets where external context drives regime changes.
Finally, extending SAGE to multivariate-to-multivariate generation and probabilistic forecasting would broaden its applicability to real-world decision-making scenarios.

\bibliography{aaai2027}
\onecolumn

\appendix

\par\noindent
{\large\bfseries Appendix A: Full Per-Horizon Forecasting Results\par}
\vspace{0.5\baselineskip}
\begin{table*}[htbp]
\centering
\caption{Full per-horizon long-term forecasting results (MSE $\downarrow$ / MAE $\downarrow$). We compare with representative methods spanning 2023--2025 under lookback $T{=}96$. \textbf{Bold}: best; \underline{underline}: second best. ``-'': not available from the source.}
\label{tab:full_results_per_horizon}
\vskip 0.1in
\resizebox{\textwidth}{!}{%
\begin{tabular}{cc|cc|cc|cc|cc|cc|cc|cc|cc|cc}
\toprule
\multirow{2}{*}{} & \multirow{2}{*}{$T_{\mathrm{pred}}$} & \multicolumn{2}{c}{DLinear} & \multicolumn{2}{c}{TimesNet} & \multicolumn{2}{c}{PatchTST} & \multicolumn{2}{c}{FredFormer} & \multicolumn{2}{c}{iTransformer} & \multicolumn{2}{c}{Amplifier} & \multicolumn{2}{c}{DUET} & \multicolumn{2}{c}{SRSNet} & \multicolumn{2}{c}{SAGE} \\
& & \multicolumn{2}{c}{\scalebox{0.8}{\citeyearpar{zeng2023transformers}}} & \multicolumn{2}{c}{\scalebox{0.8}{\citeyearpar{wu2022timesnet}}} & \multicolumn{2}{c}{\scalebox{0.8}{\citeyearpar{nie2022time}}} & \multicolumn{2}{c}{\scalebox{0.8}{\citeyearpar{piao2024fredformer}}} & \multicolumn{2}{c}{\scalebox{0.8}{\citeyearpar{liu2024itransformer}}} & \multicolumn{2}{c}{\scalebox{0.8}{\citeyearpar{fei2025amplifier}}} & \multicolumn{2}{c}{\scalebox{0.8}{\citeyearpar{qiu2025duet}}} & \multicolumn{2}{c}{\scalebox{0.8}{\citeyearpar{wu2026enhancing}}} & \multicolumn{2}{c}{\scalebox{0.8}{Ours}} \\
\cmidrule(lr){3-4}\cmidrule(lr){5-6}\cmidrule(lr){7-8}\cmidrule(lr){9-10}\cmidrule(lr){11-12}\cmidrule(lr){13-14}\cmidrule(lr){15-16}\cmidrule(lr){17-18}\cmidrule(lr){19-20}
& & \scalebox{0.85}{MSE} & \scalebox{0.85}{MAE} & \scalebox{0.85}{MSE} & \scalebox{0.85}{MAE} & \scalebox{0.85}{MSE} & \scalebox{0.85}{MAE} & \scalebox{0.85}{MSE} & \scalebox{0.85}{MAE} & \scalebox{0.85}{MSE} & \scalebox{0.85}{MAE} & \scalebox{0.85}{MSE} & \scalebox{0.85}{MAE} & \scalebox{0.85}{MSE} & \scalebox{0.85}{MAE} & \scalebox{0.85}{MSE} & \scalebox{0.85}{MAE} & \scalebox{0.85}{MSE} & \scalebox{0.85}{MAE} \\
\midrule
  \multirow{4}{*}{\rotatebox[origin=c]{90}{ETTh1}} & 96 & 0.386 & \underline{0.400} & 0.384 & \underline{0.400} & 0.414 & 0.423 & 0.378 & \underline{0.400} & 0.386 & 0.407 & \textbf{0.376} & \textbf{0.399} & \underline{0.377} & \textbf{0.399} & 0.383 & 0.402 & 0.383 & 0.406 \\
   & 192 & 0.437 & \underline{0.432} & 0.436 & \textbf{0.429} & 0.460 & 0.444 & 0.435 & 0.433 & 0.441 & \underline{0.432} & 0.442 & 0.437 & \underline{0.429} & \textbf{0.429} & 0.433 & \underline{0.432} & \textbf{0.428} & 0.437 \\
   & 336 & 0.481 & \underline{0.459} & 0.491 & 0.469 & 0.501 & 0.469 & 0.485 & 0.464 & 0.487 & 0.461 & 0.478 & \underline{0.459} & \underline{0.471} & \textbf{0.454} & 0.476 & \underline{0.459} & \textbf{0.467} & 0.461 \\
   & 720 & 0.519 & 0.506 & 0.521 & 0.500 & 0.500 & \underline{0.464} & 0.496 & 0.469 & 0.503 & 0.468 & 0.501 & 0.471 & 0.496 & 0.468 & \textbf{0.474} & \textbf{0.455} & \underline{0.483} & 0.484 \\
\midrule
  \multirow{4}{*}{\rotatebox[origin=c]{90}{ETTh2}} & 96 & 0.333 & 0.387 & 0.340 & 0.382 & 0.302 & 0.348 & \underline{0.291} & 0.347 & 0.297 & 0.349 & 0.298 & \underline{0.346} & 0.300 & 0.354 & 0.296 & 0.350 & \textbf{0.289} & \textbf{0.340} \\
   & 192 & 0.477 & 0.476 & 0.402 & 0.414 & 0.388 & 0.404 & 0.372 & 0.401 & 0.380 & 0.402 & 0.378 & \underline{0.399} & 0.372 & 0.402 & \underline{0.370} & 0.402 & \textbf{0.367} & \textbf{0.397} \\
   & 336 & 0.594 & 0.541 & 0.452 & 0.440 & 0.426 & \textbf{0.431} & 0.419 & 0.434 & 0.428 & \underline{0.432} & 0.428 & \underline{0.432} & 0.415 & \underline{0.432} & \textbf{0.413} & 0.433 & \underline{0.414} & \textbf{0.431} \\
   & 720 & 0.831 & 0.657 & 0.462 & 0.447 & 0.431 & \underline{0.432} & 0.431 & 0.438 & 0.427 & \textbf{0.431} & 0.452 & 0.444 & \textbf{0.416} & \underline{0.432} & 0.425 & 0.438 & \underline{0.423} & 0.445 \\
\midrule
  \multirow{4}{*}{\rotatebox[origin=c]{90}{ETTm1}} & 96 & 0.345 & 0.372 & 0.338 & 0.375 & 0.329 & 0.367 & 0.326 & 0.365 & 0.334 & 0.366 & 0.320 & 0.363 & 0.324 & 0.363 & \underline{0.319} & \underline{0.362} & \textbf{0.318} & \textbf{0.359} \\
   & 192 & 0.380 & 0.389 & 0.374 & 0.387 & 0.367 & \underline{0.385} & 0.363 & 0.386 & 0.377 & 0.392 & 0.364 & 0.387 & 0.369 & 0.389 & \textbf{0.359} & \underline{0.385} & \underline{0.362} & \textbf{0.382} \\
   & 336 & 0.413 & 0.413 & 0.410 & 0.411 & 0.399 & 0.410 & \underline{0.395} & 0.410 & 0.426 & 0.426 & \underline{0.395} & \underline{0.409} & 0.404 & 0.413 & \textbf{0.391} & \textbf{0.406} & 0.399 & 0.412 \\
   & 720 & 0.474 & 0.449 & 0.478 & 0.453 & \textbf{0.454} & \textbf{0.439} & \underline{0.456} & \underline{0.445} & 0.491 & 0.463 & 0.462 & 0.450 & 0.463 & 0.448 & 0.470 & 0.455 & 0.458 & 0.448 \\
\midrule
  \multirow{4}{*}{\rotatebox[origin=c]{90}{ETTm2}} & 96 & 0.193 & 0.274 & 0.187 & 0.267 & \underline{0.175} & \underline{0.259} & 0.177 & 0.262 & 0.180 & 0.264 & 0.178 & 0.262 & \textbf{0.174} & \textbf{0.258} & 0.181 & 0.264 & 0.176 & 0.268 \\
   & 192 & 0.284 & 0.339 & 0.249 & 0.311 & \textbf{0.241} & \textbf{0.305} & \underline{0.243} & \underline{0.306} & 0.250 & 0.311 & \underline{0.243} & \underline{0.306} & \underline{0.243} & \textbf{0.305} & \underline{0.243} & 0.307 & \underline{0.243} & \textbf{0.305} \\
   & 336 & 0.369 & 0.393 & 0.321 & 0.360 & 0.305 & 0.349 & \textbf{0.302} & \textbf{0.347} & 0.311 & 0.352 & 0.305 & \underline{0.348} & \underline{0.304} & \underline{0.348} & 0.306 & 0.350 & 0.308 & \textbf{0.347} \\
   & 720 & 0.554 & 0.500 & 0.408 & 0.414 & 0.402 & 0.416 & 0.404 & 0.414 & 0.412 & 0.420 & \textbf{0.393} & \textbf{0.409} & \underline{0.399} & \underline{0.413} & 0.407 & 0.416 & 0.407 & 0.416 \\
\midrule
  \multirow{4}{*}{\rotatebox[origin=c]{90}{ECL}} & 96 & 0.197 & 0.282 & 0.168 & 0.272 & 0.181 & 0.269 & 0.148 & 0.243 & 0.148 & 0.247 & 0.149 & 0.244 & \underline{0.145} & \underline{0.240} & 0.161 & 0.257 & \textbf{0.136} & \textbf{0.231} \\
   & 192 & 0.196 & 0.285 & 0.184 & 0.278 & 0.199 & 0.289 & 0.165 & 0.260 & \underline{0.162} & \underline{0.257} & 0.165 & 0.259 & 0.163 & \underline{0.257} & 0.172 & 0.265 & \textbf{0.157} & \textbf{0.249} \\
   & 336 & 0.209 & 0.296 & 0.198 & 0.296 & 0.212 & 0.300 & 0.180 & 0.274 & 0.178 & 0.275 & 0.176 & \underline{0.270} & \underline{0.175} & \underline{0.270} & 0.190 & 0.285 & \textbf{0.171} & \textbf{0.267} \\
   & 720 & 0.245 & 0.319 & 0.218 & 0.310 & 0.228 & 0.311 & 0.218 & 0.307 & 0.225 & 0.317 & \underline{0.204} & \underline{0.296} & \underline{0.204} & \underline{0.296} & 0.231 & 0.318 & \textbf{0.195} & \textbf{0.293} \\
\midrule
  \multirow{4}{*}{\rotatebox[origin=c]{90}{Traffic}} & 96 & 0.650 & 0.396 & 0.593 & 0.388 & 0.469 & 0.363 & 0.403 & 0.273 & \underline{0.399} & \underline{0.268} & 0.450 & 0.307 & 0.407 & 0.272 & 0.471 & 0.314 & \textbf{0.393} & \textbf{0.259} \\
   & 192 & 0.598 & 0.370 & 0.617 & 0.396 & 0.467 & 0.355 & 0.429 & 0.282 & \textbf{0.421} & \underline{0.276} & 0.489 & 0.325 & 0.431 & 0.284 & 0.480 & 0.316 & \underline{0.422} & \textbf{0.274} \\
   & 336 & 0.605 & 0.373 & 0.629 & 0.399 & 0.475 & 0.358 & 0.441 & 0.287 & \textbf{0.437} & \underline{0.284} & 0.484 & 0.319 & 0.456 & 0.299 & 0.496 & 0.325 & \underline{0.439} & \textbf{0.281} \\
   & 720 & 0.645 & 0.394 & 0.640 & 0.396 & 0.514 & 0.375 & \textbf{0.463} & \textbf{0.297} & 0.471 & 0.300 & 0.517 & 0.345 & 0.509 & 0.334 & 0.531 & 0.353 & \underline{0.468} & \underline{0.298} \\
\midrule
  \multirow{4}{*}{\rotatebox[origin=c]{90}{Weather}} & 96 & 0.196 & 0.247 & 0.172 & 0.214 & 0.177 & 0.228 & \underline{0.163} & \underline{0.207} & 0.174 & 0.217 & 0.165 & 0.210 & \underline{0.163} & \underline{0.207} & 0.167 & 0.211 & \textbf{0.157} & \textbf{0.201} \\
   & 192 & 0.237 & 0.267 & 0.219 & 0.260 & 0.225 & 0.265 & 0.224 & 0.260 & 0.221 & 0.261 & \underline{0.212} & \underline{0.256} & 0.218 & 0.260 & 0.215 & 0.258 & \textbf{0.206} & \textbf{0.248} \\
   & 336 & 0.283 & 0.298 & 0.280 & 0.306 & 0.278 & 0.300 & 0.278 & 0.301 & 0.278 & 0.302 & \underline{0.267} & \underline{0.296} & 0.274 & 0.299 & 0.270 & 0.298 & \textbf{0.266} & \textbf{0.292} \\
   & 720 & 0.345 & \textbf{0.335} & 0.365 & 0.359 & 0.354 & 0.348 & 0.357 & 0.348 & 0.358 & 0.352 & \underline{0.344} & 0.343 & 0.349 & 0.346 & 0.346 & 0.343 & \textbf{0.343} & \underline{0.341} \\
\midrule
  \multirow{4}{*}{\rotatebox[origin=c]{90}{Exchange}} & 96 & 0.088 & 0.218 & 0.107 & 0.234 & 0.086 & 0.215 & \underline{0.084} & \underline{0.202} & 0.086 & 0.218 & \textbf{0.083} & \underline{0.202} & 0.085 & \underline{0.202} & \underline{0.084} & \textbf{0.201} & \underline{0.084} & 0.205 \\
   & 192 & 0.176 & 0.315 & 0.226 & 0.344 & 0.205 & 0.335 & 0.183 & 0.302 & 0.177 & 0.306 & 0.175 & \underline{0.297} & 0.173 & 0.299 & \underline{0.171} & 0.303 & \textbf{0.170} & \textbf{0.296} \\
   & 336 & \textbf{0.308} & 0.415 & 0.431 & 0.487 & 0.385 & 0.459 & 0.335 & 0.418 & 0.331 & 0.422 & 0.328 & 0.414 & 0.322 & \textbf{0.408} & 0.324 & \underline{0.410} & \underline{0.315} & \underline{0.410} \\
   & 720 & 0.839 & 0.695 & 0.901 & 0.714 & 0.793 & 0.672 & 0.893 & 0.711 & 0.847 & 0.695 & 0.858 & 0.696 & \underline{0.785} & 0.688 & 0.792 & \underline{0.667} & \textbf{0.758} & \textbf{0.659} \\
\midrule
  \multicolumn{2}{c|}{\textbf{1st Count}} & 1 & 1 & 0 & 1 & 2 & 3 & 2 & 2 & 2 & 1 & 3 & 2 & 2 & 6 & 4 & 3 & 16 & 19 \\
\bottomrule
\end{tabular}%
}
\end{table*}


\end{document}